\documentclass[journal]{IEEEtran}

\usepackage{cite}
\usepackage{amsmath,amssymb,amsfonts}
\usepackage{algorithmic}
\usepackage{graphicx}
\usepackage{textcomp}
\usepackage{amssymb}
\usepackage{pifont}
\usepackage{array}
\usepackage[T1]{fontenc}

\usepackage{hyperref}

\ifCLASSINFOpdf
\else
\fi

\begin{document}
%
\title{RT-DEMT: A hybrid real-time acupoint detection model combining mamba and transformer}
%
%
%

\author{
Shilong Yang, Qi Zang, Chulong Zhang, Lingfeng Huang, Yaoqin Xie
\thanks{Shilong Yang, Chulong Zhang, Lingfeng Huang, and Yaoqin Xie are with the Shenzhen Institutes of Advanced Technology, Chinese Academy of Sciences, Shenzhen, China, 518055.}
\thanks{Qi Zang is with the Qingdao University, Qingdao, China, 266000.}
\thanks{corresponding authors: Yaoqin Xie (e-mail:yq.xie@siat.ac.cn)}
}

\markboth{}
{Shell \MakeLowercase{\textit{et al.}}: Bare Demo of IEEEtran.cls for IEEE Journals}

\maketitle

\begin{abstract}
Traditional Chinese acupuncture methods often face controversy in clinical practice due to their high subjectivity. Additionally, current intelligent-assisted acupuncture systems have two major limitations: slow acupoint localization speed and low accuracy. To address these limitations, a new method leverages the excellent inference efficiency of the state-space model Mamba, while retaining the advantages of the attention mechanism in the traditional DETR architecture, to achieve efficient global information integration and provide high-quality feature information for acupoint localization tasks. Furthermore, by employing the concept of residual likelihood estimation, it eliminates the need for complex upsampling processes, thereby accelerating the acupoint localization task. Our method achieved state-of-the-art (SOTA) accuracy on a private dataset of acupoints on the human back, with an average Euclidean distance pixel error (EPE) of 7.792 and an average time consumption of 10.05 milliseconds per localization task. Compared to the second-best algorithm, our method improved both accuracy and speed by approximately 14\%. This significant advancement not only enhances the efficacy of acupuncture treatment but also demonstrates the commercial potential of automated acupuncture robot systems.
Access to our method is available at \href{https://github.com/Sohyu1/RT-DEMT}{https://github.com/Sohyu1/RT-DEMT}
\end{abstract}


\IEEEpeerreviewmaketitle

\section{Introduction}
\IEEEPARstart{B}{reast} 
As artificial intelligence technology advances, AI has increasingly intersected with numerous fields, particularly aligning well with the medical domain. The continuous growth in demand for rapid and accurate diagnostic and treatment services has also become a key driver in the advancement of intelligent healthcare. The automation of clinical tasks with robotics and artificial intelligence technology has great potential in improving the accuracy, consistency, and accessibility of treatment \cite{1} \cite{15}. Intelligent acupuncture robot is one of the most promising development fields \cite{12} \cite{13} \cite{14}. 

Acupuncture is a kind of traditional Chinese medical practice, which has been widely recognized in modern medical circles for its effectiveness in relieving inflammation and pain. In addition, acupuncture also shows great potential in immune regulation and treatment of some mental diseases \cite{2} \cite{6} \cite{7} \cite{10}.
Related research proved the feasibility of acupuncture as an alternative therapy for inflammation and pain \cite{2} \cite{6} \cite{7}.
Moreover, some scholars have found that acupuncture  can independently mobilize the body's self-healing mechanism to restore the body's homeostasis, so as to effectively treat immune diseases \cite{3}.
In addition, some studies tried to apply acupuncture in the treatment of mental diseases, such as depression and Parkinson's disease, and achieved remarkable results, thus expanding its scope of application \cite{8} \cite{10}.
And now many scholars tend to link acupuncture and neuroscience to explain the clinical applicability of acupuncture \cite{4} \cite{5} \cite{9}.

However, one of the critical challenges in developing effective acupuncture robots is the accurate localization of acupoints. Acupoints localization tasks are often solved using techniques related to key points localization \cite{1}. In computer vision field,key point detection is a core subtask within pose estimation, involving the identification of critical human body parts such as the head, shoulders, elbows, wrists, and knees. Pose estimation is a significant research area in the field of computer vision, primarily aimed at identifying various human body parts from images or videos and accurately estimating their spatial positions and orientations. However, there are several different aspects between the acupoints detection field and point detection in CV field, which makes acupoints detection more challenging. Firstly,the number of acupoints is much more than key points in the task of pose estimation. Secondly, due to the lack of corner points, skin texture features, and other obvious features, the task of feature extraction becomes difficult.  This table, along with the acupoint positioning diagram in the upper left corner of Figure \ref{fig_mom0}, clearly outlines the method for locating acupoints based on the human skeletal structure. The relative positions of acupoints vary with different skeletal forms, adding complexity to intelligent acupoint localization tasks. More precise location standards will further advance posture estimation solutions. And, in the description of localization methods, the unit “cun” can be dynamically defined as a fraction of the distance between certain parts of the patient’s own body, and used as a unit for measuring acupoint locations. There are two main methods for determining this: the bone proportional measurement and the finger cun method. In traditional Chinese medicine clinical practice, the latter is more commonly used due to the difficulty of conveniently obtaining the distance between the patient’s bones and joints\cite{66} \cite{67}.

\begin{table*}[htbp]
\caption{Method for Locating Back Acupoints Based on the National Standard Definitions of Acupoint Names and Positions.}
\begin{tabular}{l c } 
 \hline
  acupoint & Location Method  \\ 
 \hline
  dazhui & Underspinous process depression of the 7th cervical vertebrae.  \\
  taodao & In the depression below the spinous process of the first thoracic vertebra.  \\
  shenzhu & The depression beneath the spinous process of the third thoracic vertebra  \\
  shendao & Central part of the fifth and sixth thoracic vertebrae  \\
  lingtai &  The sixth thoracic vertebrae are located in a depression beneath the spinous process \\
  zhiyang &  In the depression below the spinous process of the 7th thoracic vertebra \\
  jinsuo &  The depression beneath the ninth thoracic spine \\
  zhongshu &  10th thoracic spine spinous process depression \\
  jizhong &11th thoracic spine spinous process depression \\
  xuanshu &Lower spinous process depression of the first lumbar vertebra \\
  mingmen &Located between the spinous processes of the second and third lumbar vertebrae\\
  yaoyangguan &Lower spinous depression of the fourth lumbar spine \\
  dashu & Left and right 1.5 "cun" of spinous process of the first thoracic vertebra \\
  fufen & Left and right 3 "cun" of spinous process of the second thoracic vertebra \\ 
  ruyu & When the crease behind the armpit is straight and the lower edge of the scapula is depressed. \\
  quyu & At the midpoint of the line connecting Xiaoyu and the spinous process of the second thoracic vertebra. \\
  riyue & The gap between the 7th rib and the front midline is opened by 4 "cun" \\
  chize & In the concave groove of the elbow crease\\
  taiyuan & Ulnar depression of extensor pollicis longus tendon \\
  
\hline
\end{tabular}
\label{Table0}
\end{table*}

The initial pose estimation tasks are often solved using traditional image processing techniques, such as edge detection and shape matching. With the emergence of deep learning techniques such as convolution networks and transform networks, and their continuous application in pose estimation tasks \cite{17} \cite{18} \cite{19} \cite{21} \cite{23} \cite{25}. Now pose estimation tasks has evolved into two main methods: top-down and bottom-up approaches. The top-down approach requires first detecting the bounding boxes of all targets, and then detecting the key points within these boxes. This method usually provides higher accuracy, but it is slower and prone to errors such as missed detection and false detection in bounding box detection \cite{16}, \cite{24}, \cite{26}. On the contrary, the bottom-up approach first detects all key points and then assembles them into human body shapes, sacrificing some accuracy to improve processing speed \cite{22}, \cite{27}.

There are two approaches for identifying and locating key points: Registration-based and Detection-based.
The registration based key points recognition and localization method first aligns the same structures in different images into a common coordinate framework, and then matches and locates key points based on registration knowledge. This method first requires a predefined model that defines the structure and relative positions of key points. The purpose of registration is to minimize the error between template key points and their corresponding positions in the image \cite{32}, \cite{33}, \cite{34}, \cite{35}.
Key point recognition and localization based on detection first utilizes the feature extraction capability of object detection networks to extract image features, which are then output as key point coordinates through the classification header at the end of the network \cite{22}, \cite{37}, \cite{36}. The classification header is often divided into three technical routes: Regression-Based\cite{28} \cite{30} \cite{31} \cite{32}, Heatmap-Based\cite{20}, and Coordinate Classification (SimCC)\cite{29}.

The feature extraction capability of object detection networks determines the quality of image features and further affects the accuracy of the final localization task.
About object detection networks, YOLO, popular for its balance between speed and accuracy, is negatively impacted by Non-Maximum Suppression (NMS) \cite{39} \cite{40} \cite{41} \cite{42} \cite{43}. The Transformer based end-to-end detector (DETR), an alternative that eliminates NMS, suffers from high computational costs \cite{44}, \cite{45}. RT-DETR \cite{46}, a real-time, end-to-end object detection model with an efficient hybrid encoder, this decoder rapidly processes multi-scale features by decoupling intra-scale interactions and cross-scale fusion to enhance speed. Subsequently, utilize uncertainty-minimized query selection to provide high-quality initial queries for the decoder, thereby improving accuracy.
However, the image features downsampled by CNN inherently lose some precision and global information. When these truncated features are fed into a transformer architecture, they largely fail to fully unleash the potential of the global receptive field offered by the attention mechanism. Consequently, it is quite natural for us to embed Mamba into the network as the vision backbone, replacing the original CNN, to achieve the capability of extracting image features with linear complexity without sacrificing the global receptive field.

For classification header, in the Regression-based approach, all key points share the same feature, and the calculation of all key points is completed simultaneously, so the speed is faster than other methods, but the accuracy is often relatively low and overfitting is prone to occur during training \cite{28} \cite{30} \cite{31} \cite{32}.
The Heatmap-based approach generates corresponding Gaussian heatmaps through features, performs feature matching in the spatial dimension, and focuses more on and utilizes local features. Each key points are independently calculated. However, generating heatmaps through upsampling inevitably introduces quantization errors and requires a significant amount of computational resources \cite{20}.
The coordinate classification method (SimCC) combines the first two methods and reconstructs the key points localization task into a regression task on the horizontal and vertical axes, reducing computational complexity and eliminating quantization errors while maintaining accuracy \cite{29}.
Recent work introduction of residual likelihood estimation into coordinate regression tasks has achieved significant success, making regression based methods surpass heatmap based methods in accuracy for the first time \cite{31}. residual likelihood estimation is a statistical method used to estimate parameters of a model by focusing on the residuals, which are the differences between observed values and the values predicted by the model. This approach often involves maximizing the likelihood function of the residuals, assuming they follow a certain probability distribution. It is particularly useful in models where the residuals are assumed to be normally distributed and can provide more robust parameter estimates in the presence of model misspecification or outliers. In pose estimation, residual likelihood estimation offers significant benefits by enhancing robustness to noise, improving accuracy, and ensuring statistical efficiency. It allows for better model validation and flexibility in handling complex error structures, ultimately leading to enhanced predictive performance.
Recent work innovatively bypasses Gaussian heatmaps and directly calculates approximate maximum response points from feature maps, breaking down the barriers between regression based and heatmap based methods and enabling the model to make direct regression predictions on coordinate values \cite{30}. Differentiable Spatial to Numerical Transform (DSNT) is a technique used in computer vision, particularly in the context of pose estimation. It provides a way to convert spatial heatmaps into precise numerical coordinates in a differentiable manner, which is essential for end-to-end training of neural networks. The key idea behind DSNT is to use a soft-argmax operation to obtain the coordinates of key points from heatmaps, allowing the gradients to flow through the entire network during back-propagation. DSNT offers several advantages in pose estimation: it is fully differentiable, enabling end-to-end training within neural networks; it converts heatmaps to precise numerical coordinates, essential for high-accuracy tasks; it ensures smooth gradients through the soft-argmax operation, facilitating stable and efficient training; it is robust to noisy heatmaps by considering the entire distribution rather than just peak values; and it is easily integrable with various network architectures and heatmap-based keypoint detection methods. Thus, DSNT is a valuable tool for precise and robust keypoint localization in pose estimation.

Mamba\cite{47} \cite{53} is a novel architecture based on the state space model designed to address the limitations of traditional transformer models\cite{48} in handling long sequences efficiently. The state space model consists of two main equations: the state equation and the observation equation. This equation describes how the state of the system evolves over time. Mamba integrates selective state space models (SSMs) \cite{52}with hardware-aware algorithms, enabling linear-time sequence modeling without sacrificing performance on dense modalities like language, audio, and genomics. Observation equation describes how the observations (or measurements) are related to the state of the system. Mamba replaces the attention mechanism found in transformers with a simplified structure that combines SSMs with multi-layer perceptron (MLP) blocks. Mamba’s design ensures smooth scaling with model size, maintaining high pretraining quality and downstream performance across various tasks and datasets. By leveraging techniques like kernel fusion and recomputation, Mamba minimizes memory usage, aligning with the requirements of modern hardware accelerators like GPUs.Mamba addresses the pain points of SSM through three optimizations: discretized SSM, cyclic/convolutional representation, and HiPPO-based long sequence processing\cite{61}\cite{63}. Initially, it employs zero-order hold techniques for continuous representation and sampling. When Mamba receives a discrete signal, it retains its value until a new discrete signal is received, thereby creating a continuous signal usable by SSM. This retention period is determined by a learnable parameter, the step size (siz), which represents the resolution of the input’s phase hold. With the continuous input signal, continuous output can be generated, and values are sampled based solely on the input’s step size. Mamba can be represented in the form of a convolution, allowing for parallel training similar to Convolutional Neural Networks (CNNs). However, due to the fixed size of convolution kernels, their inference is not as fast as that of Recurrent Neural Networks (RNNs). Consequently, Mamba employs a strategy of using an RNN structure for inference and a CNN structure for training, significantly enhancing inference speed.

Recent studies have found that combining Mamba\cite{47} \cite{53} and Transformer modules\cite{48} yields significantly better results than using each individually\cite{49} \cite{50} \cite{51}. This is because the integration of Mamba’s long-sequence processing capability with the Transformer’s modeling ability can markedly enhance computational efficiency and model performance.

Our work focuses on the clinical pain points of acupuncture: precision and speed. We combined mamba and transformer to build a real-time and accurate acupoints recognition and positioning network. Our primary contribution include:

\begin{itemize}
\item[$\bullet$] We leverage the excellent long-sequence processing capability and inference efficiency of the state space model Mamba, while retaining the global modeling advantages of the attention mechanism in the traditional DETR architecture, to achieve efficient global information integration. This integration significantly enhances computational efficiency and model performance while reducing the number of parameters.
\item[$\bullet$] Our model achieved state-of-the-art accuracy and detection rate on our proprietary human back acupoint detection dataset. 
\item[$\bullet$] Our network architecture fully leverages the advantages of the state space model Mamba, which can be approximated as an RNN during inference, significantly reducing the model’s runtime. Additionally, the classification head in the network bypasses the upsampling operation in the keypoint localization step. By employing a residual likelihood estimation approach, the model enhances speed while maintaining accuracy.
\end{itemize}

\section*{Method}

Our study is aimed at developing an efficient key points detection network, the RT-DEMT, to address the challenges of speed and accuracy in acupuncture clinical settings. In this section, we described our proposed RT-DEMT network architecture. In the first subsection, we formalized the key point detection task. In the second subsection, we provided a detailed description of the RT-DEMT framework. In the third subsection, we introduced the evaluation metrics used for acupoints detection tasks.

\subsection{Key Points Detection Task}
\begin{figure*}[t]
    \centering 
    \includegraphics[width=1\textwidth, angle=0]{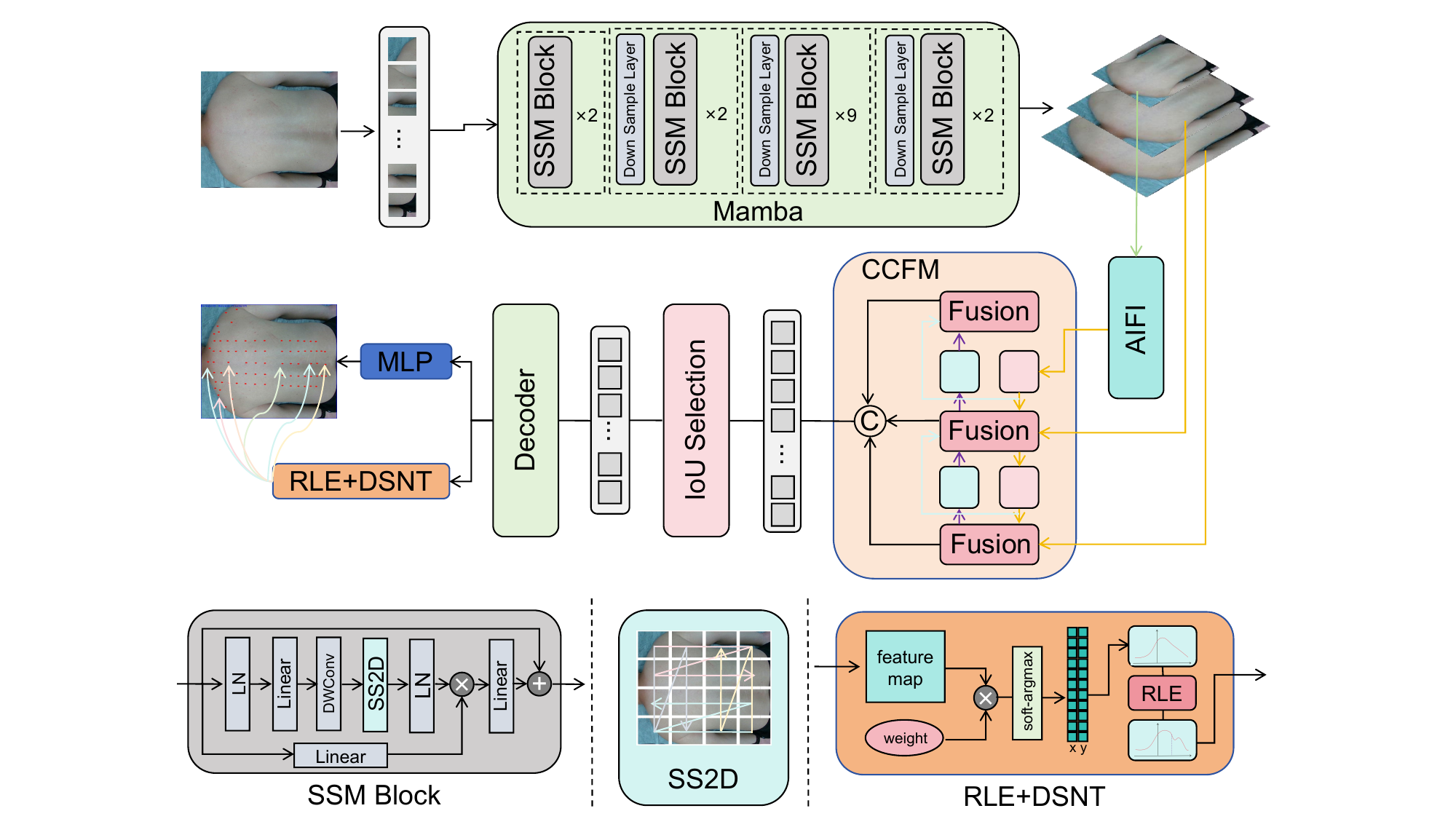}	
    \caption{our network structure}
    \label{fig_mom1}%
\end{figure*}
The key point detection task can be viewed as a regression problem, where the objective is to predict the coordinates of key points in the human body in the image. Given an image $I$ containing a human body, our goal is to predict the coordinates of $N$ key points of the human body $((x_1, y_1), \ldots, (x_N, y_N))$. We can define a mapping function $f$ that maps the image $I$ to a $2N$ dimensional output vector $\mathbf {p}$, where
\begin{equation}
     \mathbf{p} = [x_1, y_1, \ldots, x_N, y_N]^T 
\end{equation}
The problem of key points location can be formalized as the following optimization problem:
\begin{equation}
    \min \sum_{i=1}^m \mathcal{L}(\mathcal{F}(I_i), \mathbf{p}_i)
\end{equation}
Among them, $m$ is the number of training samples, $\mathcal{F}$ is a concise representation of the model, $I_i$ is the image of the $i$ training sample, and $\mathbf {p}_i$ is the corresponding real key point coordinate, $\mathcal{L}$ is the loss function used to measure the difference between predicted key points and real key points.

The $\mathcal{F}$ can be represented as follow.
\begin{equation}
    \mathcal{F}(I_i) = \mathcal{H}(\mathcal{D}(\mathcal{E}_{hy}(\textit{Mamba}(I_i))))
\end{equation}
The first component is the Mamba section, serialize the image and input it into the Mamba Backbone for feature extraction, and get ${F_{\text{py}\cdot}}_{i}$: 
The size of the extracted multi-layer features ${F_{\text{py}\cdot}}_{i}$ is as follows:
\begin{center}
    $ \begin{Bmatrix}
    &  & \frac{h}{32} &  \frac{w}{32}\\
    batch size & in channel &  \frac{h}{16} & \frac{w}{16}\\
    &  & \frac{h}{8} & \frac{w}{8}
    \end{Bmatrix}$
\end{center} 
Feed the extracted image features into the Efficient Hybrid Encoder $\mathcal{E}_{hy}$,
${F_{\text{e}\cdot}}_{i}$ represents the encoded feature data. During this period, image features undergo intra scale interaction for feature fusion, enabling the model to better understand the relationship between the target and its surrounding environment, and greatly reducing information loss. The scale of ${F_{\text{py}\cdot}}_{i}$ is consistent with that of $ {F_{\text{e}\cdot}}_{i}$.

The encoded data is decoded by the Decoder $\mathcal{D}$ and passed to the classification head, and get ${F_{\text{d}\cdot}}_{i}$.

The classification head $\mathcal{H}$ ultimately outputs the categories, coordinates, and confidence scores of the key points required for the task.
\begin{equation}
    N \times (c_n, (x_n, y_n), s_n)_{n \in N} = \mathcal{H}({F_{\text{d}\cdot}}_{i})
\end{equation}

where $c_i$ denotes the class of the $n$-th key point, $(x_n, y_n)$ represents its coordinates, $s_n$   indicates the confidence score. and  $N$ is the number of detected key points..

The loss function $\mathcal{L}$ selects the mean square error (MSE) that is highly consistent with the inter pixel error:

\begin{equation}
    \mathcal{L}(\mathbf{\hat{p}}, \mathbf{\hat{q}}) = \frac{1}{2n} \sum_{j=1}^{2n} (\hat{p_j} - \hat{q_j})^2
\end{equation}
Among them, $\mathbf{\hat{p}}$ and $\mathbf{\hat{q}}$ represent the normalized predicted key points coordinates and the normalized actual key points coordinates, respectively.
By minimizing this loss function, we can train model $\mathcal{F}$ to more accurately predict human key points in the image.

\subsection*{Network Structure}

We elucidate the RT-DEMT network architecture, as shown in Figure \ref{fig_mom1}. mapping the acupoints detection task smoothly onto a key points detection task. Our model, built upon the foundation of RT-DETR, is optimized in two main aspects: the feature extractor and the classification head.
Following serialization, the input image is processed by a Mamba-based feature extraction network, which extracts pyramid-structured feature data enriched with high-quality global information. These image features are then fed into an efficient hybrid encoder. This encoder employs intra-scale interaction and cross-scale fusion to encode the multi-scale pyramid-structured features and utilizes an IOU querying mechanism for selection. The selected features are subsequently directed to various classification heads at the model’s end, yielding bounding boxes and key points coordinates with associated confidence scores.

In terms of the feature extractor, to fully leverage the potential of the global receptive field offered by attention mechanisms, we replace the conventional CNN backbone with Mamba based on state space model.This substitution addresses the issue of truncated features typical in CNN. Mamba enhances the efficiency and flexibility of processing sequential data through the adoption of a state-space model with time-varying parameters and the introduction of hardware-aware algorithms. It is naturally suited for handling long sequence problems. The inherent global receptive field of the state-space model-based Mamba is particularly beneficial in acupoints detection tasks, which often lack corner points and skin texture. Studies indicate that a hybrid use of Mamba with Transformer modules, combining Mamba’s long-sequence processing capabilities with the modeling power of Transformers, significantly improves computational efficiency and model performance over using either alone.

\subsubsection*{Mamba backbone}
The essence of Mamba is the state space model which represents a system using a set of first-order differential (or difference) equations\cite{52} \cite{62} \cite{64} \cite{65}. These equations describe the dynamics of the system’s state variables and how they are influenced by inputs, outputs, and noise.


We define\(\mathbf{x}_t\) is the state vector at time $t$,
\(\mathbf{A}\) is the state transition matrix,
\(\mathbf{u}_t\) is the control input vector at time $t$,
\(\mathbf{B}\) is the input matrix,
\(\mathbf{w}_t\) is the process noise vector, often assumed to be normally distributed with zero mean and covariance matrix \(\mathbf{Q}\).



The state space model can be compactly represented as:
\[
\mathbf{x}_{t+1} = \mathbf{A} \mathbf{x}_t + \mathbf{B} \mathbf{u}_t + \mathbf{w}_t \
\mathbf{y}_t = \mathbf{C} \mathbf{x}_t + \mathbf{D} \mathbf{u}_t + \mathbf{v}_t
\]

The matrices \(\mathbf{A}\), \(\mathbf{B}\), \(\mathbf{C}\), and \(\mathbf{D}\) define the system dynamics and observation model, while \(\mathbf{w}_t\) and \(\mathbf{v}_t\) represent the process and observation noise, respectively. These models are widely used because they provide a structured way to model complex systems and can be analyzed and controlled using various mathematical and computational techniques.


\subsubsection*{Classification Header}

Our classification header framework integrates the advantages of both of the RLE and DSNT, eliminating the upsampling process to reduce computational load, and introduces residual likelihood estimation to enhance the accuracy of localization.

The high-quality feature maps extracted by the combined Mamba and Transformer architecture directly compute an approximated maximum response point to obtain the distribution of x and y coordinates. Subsequently, residual likelihood estimation is introduced to perform standardized regression on this distribution. Not only do we bypass the upsampling step, significantly reducing the model’s computational complexity, but the standardization through residual likelihood estimation also achieves coordinate precision comparable to heatmap-based approaches.

\subsubsection*{Efficient Hybrid Encoder}
We can regard the Mamba architecture as a specialized CNN network, which can transform into an RNN during the inference process to achieve rapid inference. And the efficient hybrid encoder can combine the strengths of convolutional neural networks (CNNs) and Transformer architectures to effectively capture both local and global features.

The hybrid feature representation combines the local features from the convolutional layer and the global features from the Transformer encoder:
\[
\mathbf{F}_{\text{hybrid}} = \sum_{k = 1}^{pl} \mathbf{Fusion}(F_{pyramid})_{k} 
\]
where $pl$ is the total number of layers in the pyramid.

The features extracted by the Mamba architecture inherently carry global information. This combination ensures that the encoder captures both local and global context more effectively.

The IOU selection mechanism is used to evaluate the overlap between predicted bounding boxes and ground truth bounding boxes, thereby optimizing detection performance. IOU (Intersection over Union) is a commonly used metric, defined as the ratio of the area of intersection to the area of union between the predicted bounding box and the ground truth bounding box.
Specifically, assuming the predicted bounding box is \( B_p \) and the ground truth bounding box is \( B_g \), the Intersection over Union (IOU) can be expressed as:

\[
\text{IOU}(B_p, B_g) = \frac{|B_p \cap B_g|}{|B_p \cup B_g|}
\]

where \( |B_p \cap B_g| \) represents the area of the intersection between the predicted and ground truth bounding boxes, and \( |B_p \cup B_g| \) represents the area of their union.

Through this mechanism, model can effectively evaluate and optimize detection results, enhancing the model’s accuracy and robustness.
\subsection{evaluating indicator}
\begin{table*}[htbp]
\centering
\caption{Evaluate different models on the same test set for acupoint localization tasks and compare their performance.Two metrics describes the accuracy of the model’s keypoint task, include the Euclidean Pixel Error (EPE) between predicted and actual acupoint positions, and the Percentage of Correct Keypoints (PCK) with reprojection error thresholds set at 0.05 and 0.1. Additionally, model quality is described by:Parameters (M),FLOP (G),Average Inference Time (ms) ,Final Throughput.}
\begin{tabular}{l c c c c c c c} 
 \hline
  &epe & $PCK_{0.05}$ & $PCK_{0.1}$& Params(M) & Flops(G) & Average inference time(ms) & Final Throughput \\ 
 \hline
 PVT\cite{55}  & $10.64 \pm 0.2$ & 0.81  & 0.96&28&4.919 & $13.99 \pm 2$ & 66.89 \\
 
 Swim-pose\cite{56}& $11.81 \pm 0.2$ & 0.75  & 0.98&69&92.2 & $16.69\pm 2 $  & 59.27 \\
 
 Yolov8   & $14.87 \pm 0.2$ & 0.87  & 0.96&48&168.6& $13.0\pm 2 $ & 76.92 \\ 
 
 Yolov9\cite{43}  & $16.75 \pm 0.2$ & 0.93  & 0.95& \textbf{20} & 76.3&$12.3\pm 2 $ & 81.30 \\ 
 
 VitPose\cite{54}  & $10.01 \pm 0.2$ & 0.89  & 1&101&106.65& $14.42\pm 2 $ & 69.34 \\
 
 RT-DETR\cite{46}  & $12.50 \pm 0.2$ & 0.93  & 1& 68 & 136.7 & $15.97\pm 2 $ & 62.61 \\
 
 UniFormer\cite{57}& $9.09  \pm 0.2$ & 0.94  & 1& 21 & 18.36 & $11.66\pm 2 $ & 85.76 \\
 
 \textbf{ours}  & $\textbf{7.79} \pm 0.2$ & \textbf{0.97}  & \textbf{1}  &77&\textbf{15.19}&$\textbf{10.05}\pm 2 $& \textbf{99.50}	 \\ 
 \hline
\end{tabular}
\label{Table1}
\end{table*}

To quantitatively assess the performance of our key points detection algorithm, we employ the following evaluation metrics:
\begin{enumerate}
    \item \textbf{Euclidean-distance pixel error (EPE)}: The average pixel Euclidean distance between the detected key points and the ground truth key points, assessing the spatial accuracy of the detection task.
    \[\text{EPE} = \frac{1}{N} \sum_{n=1}^{N} \sqrt{(x_n - x_n^{gt})^2 + (y_n - y_n^{gt})^2}\]
     where $(x_n, y_n)$ and $(x_n^{gt}, y_n^{gt})$ are the pixel coordinates of the detected and ground truth key points, respectively.
    \item \textbf{Percentage of Correct Key points (PCK)}: Often used in human pose estimation, PCK measures the percentage of detected key points that fall within a specified distance threshold from the true key points locations. This metric is adaptable to various scales by adjusting the threshold. Here we select threshold values of 0.05 and 0.1.
    \[\text{PCK} = \frac{1}{N} \sum_{n=1}^{N} \left(\sqrt{(x_n-x^{gt}_n)^2 + (y_n-y^{gt}_n)^2} \leq \alpha \cdot \text{max}(h, w)\right)\]
    where $\alpha$ is a threshold parameter, $h$ and $w$ are the height and width of the bounding box or image.
    \item \textbf{Average inference time and Final Throughput}: The average time taken by the algorithm to process an image and detect key points, which is crucial for real-time applications. Final Throughput is the rate at which a model can process inputs over a given period, usually measured in inputs per second.
    The relationship between average inference time \((T_{\text{avg}})\) and final throughput \((F_{\text{throughput}})\) can be expressed as:
    \[ F_{\text{throughput}} = \frac{1}{T_{\text{avg}}} \]
    We can evaluate the efficiency of algorithmic reasoning from different perspectives using these two metrics.
\end{enumerate}

These metrics collectively provide a robust framework for evaluating the effectiveness and efficiency of our key points detection method.

\section*{Result}
\begin{figure*}[t]
	\centering 
	\includegraphics[width=1\textwidth, angle=0]{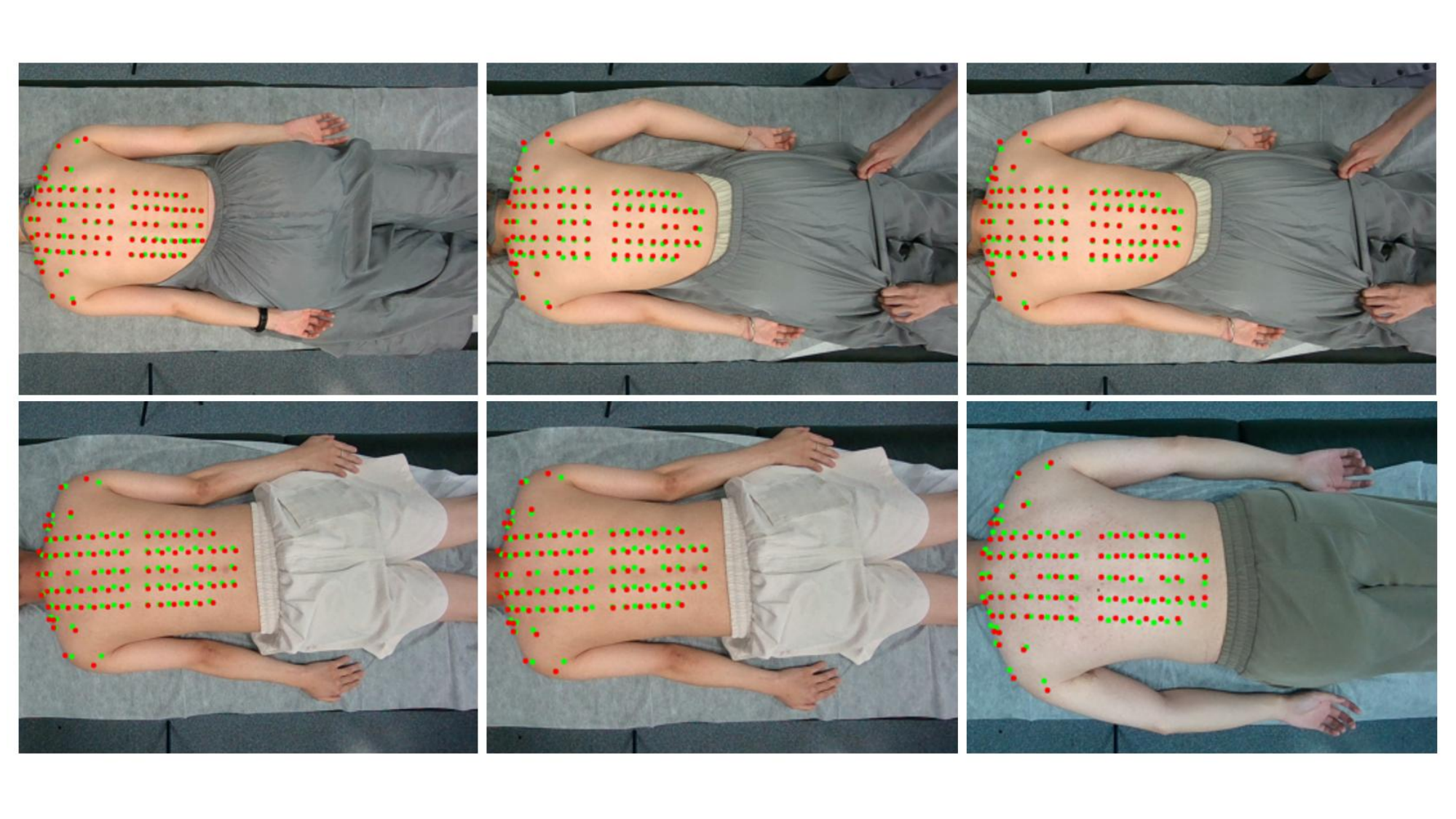}	
	\caption{The green dots represent the gold standard locations of acupoints in the test set, while the red dots indicate the predicted positions of acupoints by the model.} 
	\label{fig_mom2}%
\end{figure*}

\subsection*{Datasets and processing}
Our experimental data is based on a privately developed dataset of back acupoint locations, independently created by our team in collaboration with traditional Chinese medicine practitioners from Shenzhen People’s Hospital. Professional TCM clinicians manually annotated the 84 acupoint locations, while we assisted in the data collection process and managed it, ultimately preprocessing the experimental data to obtain a reliable clinical acupoint location dataset. We recruited nearly 200 healthy participants with normal body types and no apparent spinal deformities, all of whom provided informed consent.

\subsection*{Prediction Result}
Our network outputs a set of coordinate values and confidence scores for each acupoint prediction. By setting a confidence threshold, we filter the final outputs.
Figure \ref{fig_mom2} visually demonstrates the potential of our network in the task of acupoints localization on the human back. In Figure 2, green markers represent the ground truth distribution of acupoints, while red markers indicate the predicted positions by our network.

We compared our network with the SOAT model on the test set. We used all models to perform acupoints localization tasks on the same test set, recording the Euclidean pixel distance error (EPE) between the predicted acupoints position obtained by each model and the actual position of the acupoint, as well as the percentage of correctly located key points (PCK) with a set reprojection error threshold of 0.05 and 0.1, respectively.We also compared the networks by their parameter count, i.e., the number of parameters in the neural network (measured in millions), to represent their complexity and capacity. The computational cost of a single forward pass is measured in floating-point operations per second (FLOPs). Additionally, the average inference time is the mean time required for the network to process a single input and produce an output (measured in milliseconds), while the final throughput indicates the number of inferences the network can perform per second, reflecting its efficiency and speed in real-time applications. Combined analysis of these metrics provides an intuitive measure of the model’s real-time performance.

Table \ref{Table1} clearly demonstrates that our network has achieved a significant accuracy advantage while maintaining real-time performance. In contrast, the YOLO series models \cite{39} \cite{40} \cite{41} \cite{42} \cite{43} exhibit advantages in inference speed but fail to address localization accuracy issues. Furthermore, our method shows dual improvements in both accuracy and speed compared to RT-DETR \cite{46}, achieving nearly 1.5 times the accuracy while retaining real-time capabilities. Compared to VitPose\cite{54}, another approach utilizing transformers in the visual domain, our method also surpasses in both accuracy and speed. Additionally, VitPose\cite{54} face training difficulties due to their larger parameter sizes, which hinder their performance on smaller datasets. Uniformer ranks highly in all metrics except $pck_{0.05}$, indicating that the model’s robustness is suboptimal for tasks involving weak feature images. However, it still has significant advantages compared to other lightweight models like YOLO and PVT.Swin-pose\cite{56} also exhibits a similarly low $pck_{0.05}$ performance.

\subsection*{Alation studies}
\subsubsection*{Feature extractor}
\begin{table*}[ht]
\centering
\caption{
a comparative analysis of different feature extractor architectures based on three metrics: epe, Params (M), and ait (ms).
}
\begin{tabular}{l c c c} 
 \hline
    & epe  & Params(M) & ait(ms) \\ 
 \hline
 ResNet50\cite{58} & $9.28 \pm 0.2$ & 71  & $14.49\pm 2 $ \\
 HRNet\cite{59}     & $10.68 \pm 0.2$ & 84 & $18.64\pm 2 $ \\
 MobileNetv3\cite{60} &$11.88\pm 0.2$ & \textbf{50}  & $11.78\pm 2 $ \\
 \textbf{ours} & $\textbf{7.792}\pm 0.2$ &77&\textbf{$\textbf{10.05}\pm 2 $} \\ 
 \hline
\end{tabular}
\label{Table2}
\end{table*}
We validated the feasibility of the Mamba-based feature extractor backbone by comparing  with those of ResNet\cite{58}, HRNet\cite{59}, and MobileNetv3\cite{60} in terms of accuracy and efficiency.
In the ablation experiments of the feature extractor, we uniformly set the classification head to our proposed classification head. From  table \ref{Table2}, we can observe that our method achieved an absolute advantage in accuracy, being the only method with an average Euclidean pixel error of less than 10 on the test set, nearly doubling the accuracy of HRNet.This experiment validated that the combination of the inherent global feature extraction capability of the mamba-based feature extractor and the modeling ability of the transformer significantly enhances the model’s performance on weak feature images.

\subsubsection*{Classification Header}
\begin{table*}[ht]
\centering
\caption{
a comparative analysis of different Classification Header architectures based on three metrics: epe, Params (M), and ait (ms).
}
\begin{tabular}{l c c c} 
 \hline
    & epe  & Params(M) & ait(ms) \\ 
 \hline
 regression\cite{28} & $13.75 \pm 0.2$ & \textbf{70}  & $16.23\pm 2$ \\
 simcc\cite{29}      & $11.82 \pm 0.1$ & 78  & $24.32\pm 2$ \\
 heatmap\cite{20}    & $8.62 \pm 0.1$ & 112  & $32.58\pm 2$ \\
 \textbf{ours} &$\textbf{7.792}\pm0.1$&77 & $\textbf{10.05}\pm 2 $ \\ 
 \hline
\end{tabular}
\label{Table3}
\end{table*}
In the ablation experiments on classification heads, all models only replaced the classification head.We select a representative from each of the three mainstream classification head approaches for comparison. In table \ref{Table3} we can clearly draw the conclusion, our improved method maintains the inference speed of regression-based approaches while slightly surpassing heatmap-based methods in accuracy. This experiment validates that our classification head method achieves optimal performance in both aspects.

\section*{Discussion}
The prospect of integrating intelligence into acupuncture practices presents a transformative opportunity to enhance both the precision and efficiency of this ancient therapeutic modality. The integration of advanced technologies such as the Mamba and Transformer-based keypoint detection networks signifies a pivotal advancement in addressing the critical challenges faced in clinical acupuncture, particularly in the accurate and swift localization of acupoints.

Acupoint localization, a fundamental step in acupuncture, has traditionally been slow and imprecise, largely due to the reliance on manual identification by practitioners. This process is complicated by the vast number of acupoints and their specific anatomical correlations to bones, joints, and skin features. The task of acupoint localization aligns with the core task in pose estimation: the task of keypoint recognition, sharing a common logical framework for resolution. By leveraging such technologies, practitioners can achieve a more accurate mapping of acupoints, thereby enhancing treatment efficacy and patient outcomes.

\begin{figure*}[t]
    \centering 
    \includegraphics[width=1\textwidth, angle=0]{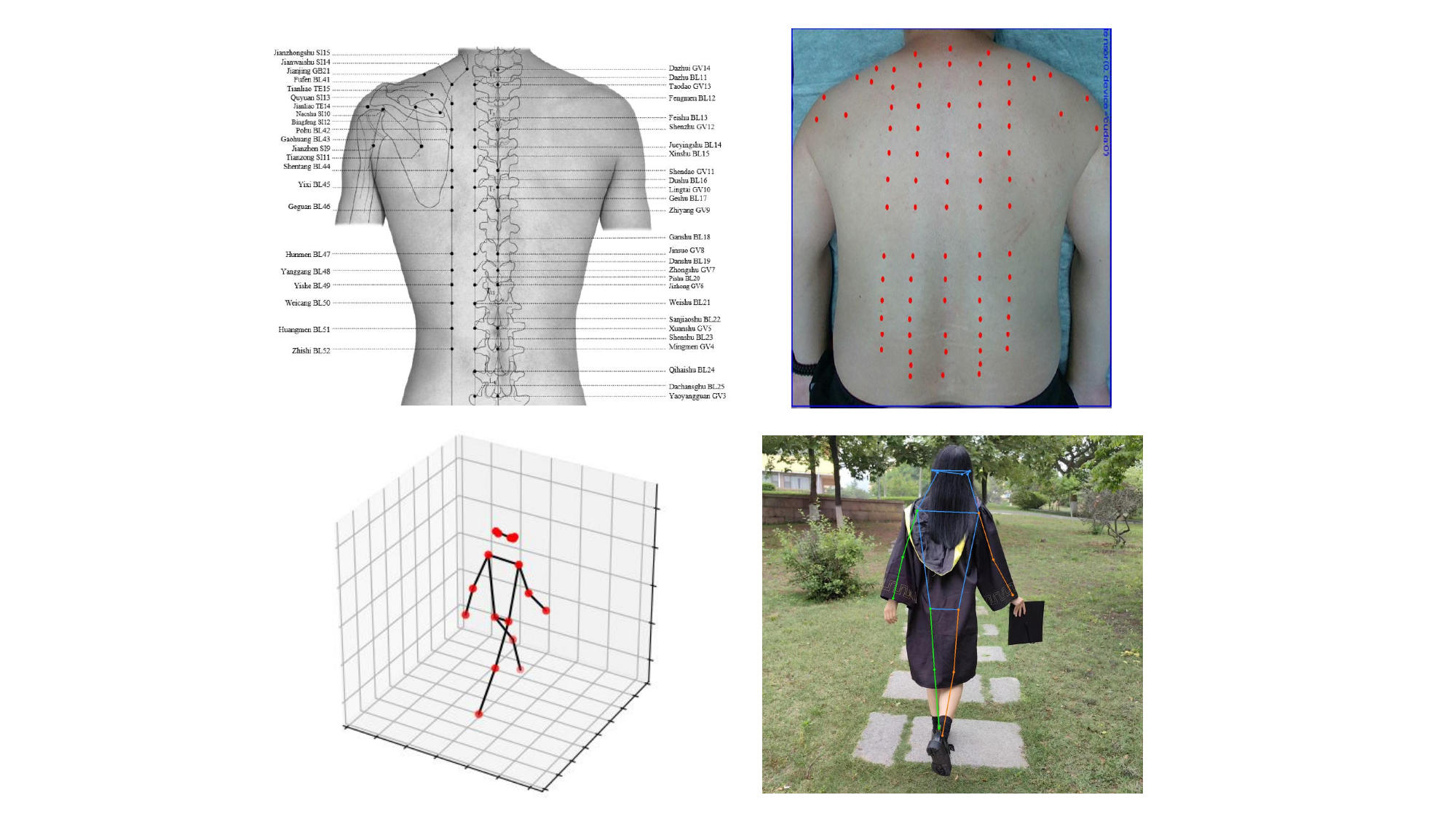}	
    \caption{The top-left image shows the standard acupoint locations on the back according to national standards, while the top-right image depicts the predicted acupoint locations from our model. The two images below represent traditional pose estimation tasks.The figure clearly shows that the number of points detected for acupoints significantly exceeds that for posture detection. This not only indicates the complexity of our task but also highlights the potential for future applications in fields such as VR and digital humans.} 
    \label{fig_mom0}%
\end{figure*}

The advancements in acupoints recognition tasks not only improve the accuracy of acupoints localization but also contribute to the broader field of posture estimation. As these technologies evolve, they enable a more nuanced understanding of human anatomy in real-time, which is crucial for both targeted acupuncture treatments and various other medical and therapeutic applications.
Moreover, in traditional pose estimation tasks, the selection of key points is often limited to a few locations such as the head, shoulders, and joints. For example, the two images in the bottom left and bottom right of Figure \ref{fig_mom0} represent solutions in traditional pose estimation tasks. However, the abundance of acupuncture points significantly expands our options, enabling more diverse pose localization.

Increasing the number of reference points and enhancing the precision of these points in localization schemes can significantly advance pose estimation solutions. More reference points provide a denser spatial framework, improving the granularity and accuracy of the estimated pose. Stricter localization criteria ensure higher fidelity in the reference data, reducing errors and ambiguities in the pose estimation process. Together, these improvements can lead to more robust and reliable pose estimation, facilitating advancements in applications such as robotics, augmented reality, and autonomous navigation.

In conclusion, the intelligent automation of acupoints detection through sophisticated neural networks like Mamba and Transformer represents a significant leap forward in the field of acupuncture. This innovation not only promises to elevate the precision and speed of treatments but also paves the way for broader applications in healthcare diagnostics and personalized medicine, ultimately leading to improved patient care and treatment outcomes.

\section*{Conclusions}
In this study, we developed a novel network by integrating Mamba and Transformer to enhance the accuracy and speed of key points recognition and localization. The network has proven effective in addressing typical weak image feature acupoint detection tasks, improving computational efficiency and detection accuracy with fewer parameters. This work not only advances the intelligentization of clinical acupuncture but also lays the foundation for more accurate and diverse posture estimation applications. We conducted extensive experiments to validate the performance of our proposed network. The results demonstrated significant improvements in both speed and accuracy compared to existing methods. Specifically, our network achieved a higher detection rate with fewer false positives, making it highly reliable for clinical applications where precision is paramount.

Beyond clinical acupuncture, the principles and techniques developed in this study have broader implications. The enhanced key point recognition and localization capabilities can be applied to various fields such as sports science, rehabilitation, and human-computer interaction. For instance, accurate posture estimation can lead to better injury prevention strategies in athletes or more effective physical therapy regimens for patients.
In conclusion, our work represents a significant step forward in the integration of advanced neural network architectures for practical applications. By combining the efficiency of Mamba with the contextual understanding of Transformers, we have created a powerful tool that not only meets the demands of clinical acupuncture but also sets the stage for future innovations in posture estimation and beyond.

\bibliographystyle{ieeetr}
\bibliography{ref}

\end{document}